\pdfoutput=1

\documentclass[11pt]{article}

\usepackage[]{acl}

\usepackage{times}
\usepackage{latexsym}
\usepackage{graphicx}
\usepackage{amsmath}
\usepackage{bbm}

\usepackage{multirow, booktabs}
\usepackage{caption}
\usepackage{subcaption}
\usepackage{makecell}

\usepackage{geometry}
\usepackage{array}
\usepackage{hhline}
\usepackage{float}
\usepackage{bm}
\usepackage{comment}

\usepackage[T1]{fontenc}
\usepackage{color, colortbl}

\usepackage[utf8]{inputenc}

\usepackage{microtype}

\definecolor{Gray}{gray}{0.92}
\newcommand{\std}[1]{\scriptsize{$\pm$#1}}
\newcommand{\argmax}[1]{\underset{#1}{\text{argmax}} \;}

\title{Teacher-Student Training for Debiasing: General Permutation Debiasing for Large Language Models}

\author{Adian Liusie, Yassir Fathullah, Mark J. F. Gales \\
  ALTA Institute, Department of Engineering, University of Cambridge \\
  \texttt{al826@cam.ac.uk, yf286@cam.ac.uk, mjfg@eng.cam.ac.uk} \\}
  
\begin{document}

\maketitle

\begin{abstract}
Large Language Models (LLMs) have demonstrated impressive zero-shot capabilities and versatility in NLP tasks, however they sometimes fail to maintain crucial invariances for specific tasks. One example is permutation sensitivity, where LLMs' outputs may significantly vary depending on the order of the input options. While debiasing techniques can mitigate these issues, and yield better performance and reliability, they often come with a high computational cost at inference. This paper addresses this inefficiency at inference time. The aim is to distill the capabilities of a computationally intensive, debiased, teacher model into a more compact student model. We explore two variants of student models: one based on pure distillation, and the other on an error-correction approach for more complex tasks, where the student corrects a single biased decision from the teacher to achieve a debiased output. Our approach is general and can be applied to both black-box and white-box LLMs. Furthermore, we demonstrate that our compact, encoder-only student models can outperform their larger, biased teacher counterparts, achieving better results with significantly fewer parameters.
\end{abstract}

\section{Introduction}

Recent advancements in Large Language Models (LLMs) have led to dramatic shifts within natural language processing (NLP). Unlike prior "pre-train and fine-tune" \cite{devlin-etal-2019-bert, he2020deberta} approaches, instruction-tuned LLMs combined with effective good prompting techniques has enabled LLMs to excel at unseen tasks without task-specific training \cite{brown2020language, touvron2023llama}. This has led to the current capabilities of LLMs, where they demonstrate great versatility, while also being powerful and displaying state-of-the-art performance on many standard NLP benchmark leaderboards \cite{Open-LLM-Leaderboard-Report-2023}.

Despite their impressive general abilities, LLMs suffer from particular limitations. They are prone to hallucinating information \cite{huang2023survey, manakul2023selfcheckgpt}, can have large sensitivity to the form of prompts \cite{sclar2023quantifying, zhou2022large} and also demonstrate systematic biases such as gender bias \cite{kotek2023gender}. Furthermore, due to the general nature of their pre-training and instruction-tuning \cite{wei2021finetuned, ouyang2022training}, for certain applications, they may be unaware of particular important task invariances. One such invariance that LLMs may fail to maintain is permutation-invariance. Ongoing work has demonstrated that LLMs can be sensitive to the input order of options, which has been observed for both question answering  \cite{pezeshkpour2023large, zheng2023large} and pairwise assessment \cite{zheng2023judging, wang2023large, liusie2023zero}. For these tasks, varying the ordering of the input options may lead to different decisions by the LLM, which can impact downstream performance and reliability.

Although debiasing approaches can be applied to enforce invariances, such methods can be computationally expensive or inapplicable to black-box settings \cite{zheng2023large}. To address these challenges, this work introduces a general framework that can be used to adapt both black-box and white-box systems to follow a particular invariance, while being inference efficient. For a given invariance and debiasing scheme, our framework trains a compact student to emulate the debiased teacher, which during inference can be efficiently deployed. We investigate two variants of students, a simple knowledge-distilled student, as well as an error-correction student that takes in a single biased teacher sample and corrects it to learn the debiased teacher decision, applicable for more complicated tasks. We demonstrate the effectiveness of our framework on permutation invariance, and illustrate that small 330M parameter student models can outperform their larger biased teacher counterparts, while also maintaining particular embedded invariances. 

The contributions of this work are: 1) We provide metrics for assessing the sensitivity of models to the input ordering of options. 2) We showcase that LLMs can demonstrate large permutation sensitivity and that biases seem correlated to task performance. 3) We study several different debiased approaches that yield significant performance gains. 4) Experiments on RACE++ and SummEval demonstrate that the teacher-student training for debiasing framework yields effective students that perform better than their biased teacher while being inference-efficient and not expensive to train.

\section{Multiple Choice Prompting}
\begin{figure*}[t]
    \centering
    \includegraphics[width=0.9\linewidth]{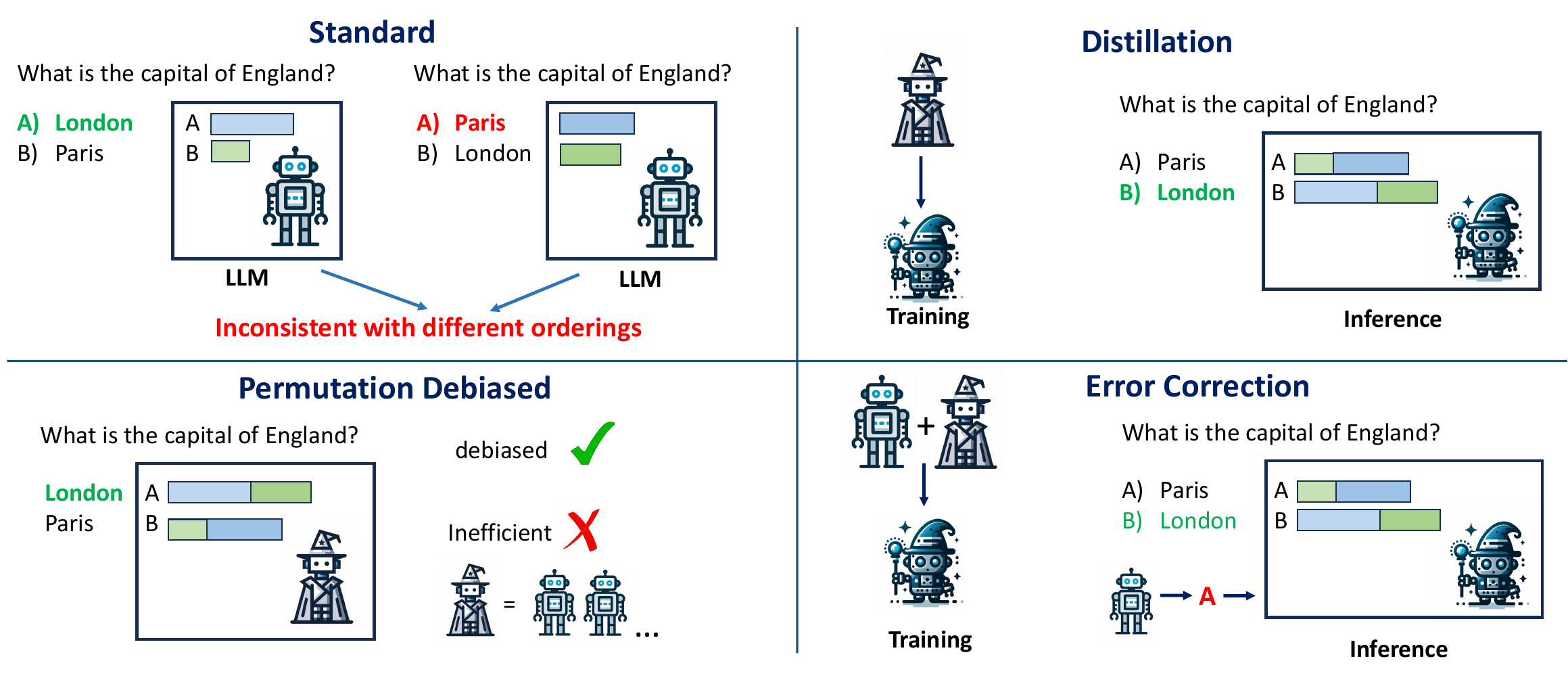}
    \caption{High-level diagram of the work: The left-hand side illustrates how LLMs may be sensitive to input ordering, but by averaging results from different permutations can yield debiased distributions. The right-hand side shows two variants of students that emulate the debiased teacher distribution, either through distillation or through error correction where the student improves a single sampled biased decision.}
    \label{fig:main-diagram}
\end{figure*}
Prompting has been shown effective in leveraging the diverse zero-shot abilities of instruction-tuned Large Language Models (LLMs). For a particular task, inputs can be rephrased into natural language queries which condition the LLM to generate useful responses \cite{reynolds2021prompt, chae2023large}. 

For example in multiple choice classification tasks, there may be input information $x$ (e.g. context and question), and a set of $K$ possible answers, $\mathcal{A}=\{a_1, ..., a_K\}$. 
Given an ordered realisation of the possible answers $\mathrm{A}_{\sigma}$ (e.g. $(a_2, a_4, a_1, a_3)$), a prompt ${\tt t}(x, \mathrm{A}_{\sigma})$ can be designed to convert the input information into a textual representation. Note that a different ordering of answers $\mathrm{A}_{\sigma_j}$, would lead to a different textual prompt representation.
The work of \citet{robinson2023leveraging} conducted a systematic study into \textit{Multiple Choice Prompting} (MCP) in which the position of each ordered answer in $\mathrm{A}_{\sigma}$ is bound to a symbol or option label $w_{k}$ within the prompt.
Instead of tasking the LLM with generating the full correct answer $a^*$ \cite{lieber2021jurassic, brown2020language}, it only needs to predict the label $w^*$ of the correct position (often a single token such as "A", "B", "C" or "D"). This converts the significant problems that arise when comparing the probabilities of variable-length answers $a \in \mathcal{A}$ into simple probabilities of different tokens.
%
%
The predictive probability of the $k$-th option under the particular permutation of answers then becomes:
\begin{gather}
    {\tt P}\left(w_k \vert x, \mathrm{A}_{\sigma}\right) \!=\! \frac{{\tt P}_{\tt LLM}(w_k|{\tt t}(x, \mathrm{A}_{\sigma}))}{\sum_{j} \! {\tt P}_{\tt LLM}(w_j \vert {\tt t}(x, \mathrm{A}_{\sigma}))}
\end{gather}
%
%
%
%
where the model probabilities have been normalized, since the LLM vocabulary span tokens beyond the symbols $w_{1:K}$. 
The probability of an answer $a$ can then be found by matching it to its corresponding position, yielding a distribution over answers ${\tt P}\left(a_k \vert x, \mathrm{A}_{\sigma}\right)$.
%
%
Overall, the system decision is the answer with the highest probability.
\begin{gather}
    \hat{a} = \argmax{a_k} {\tt P}\left(a_k \vert x, \mathrm{A}_{\sigma}\right)
\end{gather}

\noindent The above approach assumes full access to the output probabilities, which may not be available. For black-box LLMs that are served through APIs \cite{achiam2023gpt, team2023gemini}, one may only have access to the autoregressively generated output text. In such settings, one can instead randomly sample from the underlying distribution to get an approximate system decision $\tilde{a}$:
\begin{gather}
    \tilde{a}^{(i)} \!\sim {\tt P}\left(a_k \vert x, \mathrm{A}_{\sigma}\right)
\end{gather}
For well-designed prompts, the majority of the probability mass should be associated with the option labels $w_{1:K}$. One can therefore directly sample from $\tilde{w}^{(i)} \! \sim {\tt P}_{\tt LLM}(w_k|{\tt t}(x, \mathrm{A}_{\sigma}))$ and reject samples that do not belong to the options labels. 

\subsection{Multiple Choice Question Answering}
\begin{figure}[H]
    \centering
    \includegraphics[width=\linewidth]{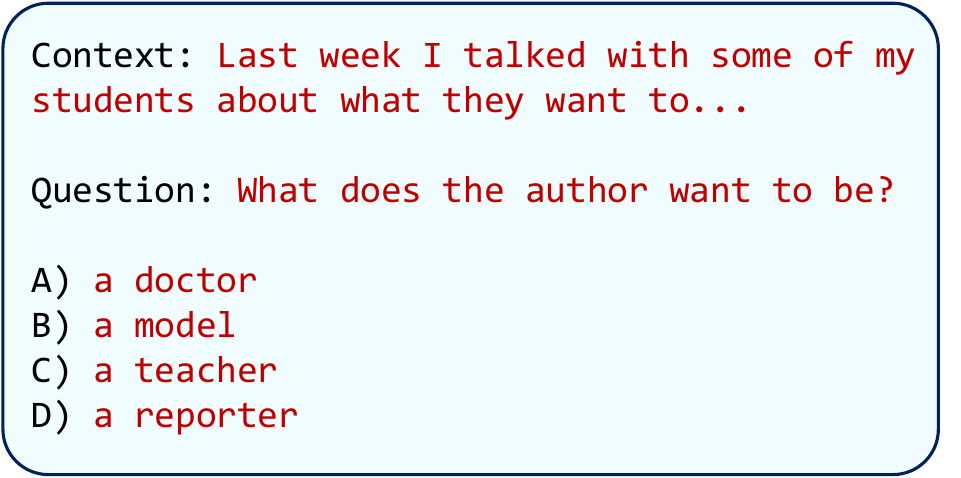}
    \caption{Templates used for prompting LLMs for  MCQA. For context-free questions, the context is omitted.}
    \label{fig:mcqa-prompt}
\end{figure}

The objective for multiple choice question answering is for a model to determine which of the provided options is the correct answer for the specified question. To determine the answer, the model must either leverage general knowledge learned in training or, if contextual information is provided, infer answers from the passage. In this work, simple prompts are used as demonstrated in Figure \ref{fig:mcqa-prompt}. 

\subsection{Comparative Asessment}
\begin{figure}[H]
    \centering
    \includegraphics[width=0.9\linewidth]{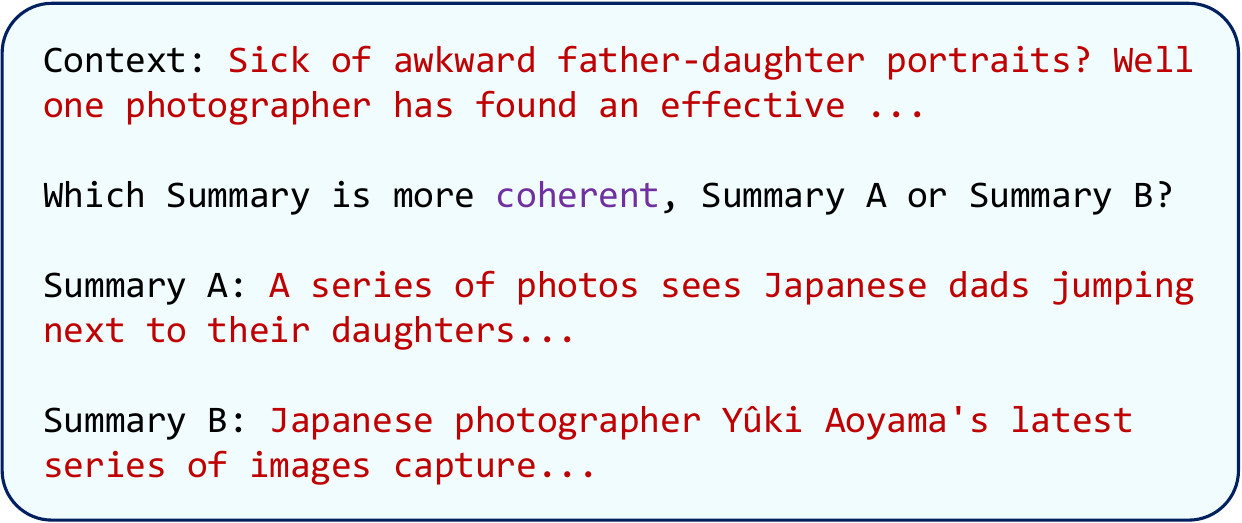}
    \caption{Prompts used for comparative assessment. Different attributes use different adjectives.}
    \label{fig:comparative-prompt}
\end{figure}

Comparative Assessment aims to determine which of two responses is better. Given a context (e.g. previous dialogue/article) the LLM is asked to assess which response is better, A or B. Comparative assessment can be used for various NLG metrics, and the prompt can be adapted towards particular attributes. The prompt used is shown in Figure \ref{fig:comparative-prompt}.

\section{Inherent Biases in LLMs}

%

%


\subsection{Quantifying Bias to Permutations}
\label{sec:bias_metrics}

Although LLMs have shown effective zero-shot performance, they may exhibit bias where they fail to recognize specific task-related invariances \cite{pezeshkpour2023large, miceli-barone-etal-2023-larger}. 
Previous work has highlighted sensitivity to the permutations of options in multiple choice question answering \cite{pezeshkpour2023large, zheng2023large}. Similarly, comparative assessment systems have been shown to favour options in particular positions \cite{zheng2023judging, wang2023large, liusie2023zero}. 
%

%
%
If a system demonstrates perfect awareness of this permutation invariance, then for any two permutations of the options, $\mathrm{A}_{\sigma_j}$, $\mathrm{A}_{\sigma_m}$, one would expect consistent distributions,
\begin{equation}
    \label{eq:unbiased}
    {\tt P}\hspace{-0.6mm}\left(a_k \vert x, \mathrm{A}_{\sigma_j}\right) = 
    {\tt P}\hspace{-0.6mm}\left(a_k \vert x, \mathrm{A}_{\sigma_m}\right)
    \hspace{2.0mm} \forall j, m
\end{equation}
%
I.e, the probability of an answer should be independent of how the options have been presented. However, the predictive distribution produced by an LLM may not conform to Equation \ref{eq:unbiased} and suffer from inherent bias, where different permutations lead to different predictive distributions. This may impact system performance, and yield biased decisions in downstream applications. To assess the sensitivity of a system to permutation, we define two metrics that can be used to measure a system's inherent bias towards a particular task.


\vspace{2mm}\noindent \textbf{Permutation Sensitivity}. As defined by Equation \ref{eq:unbiased}, the distribution over possible answers should be unaffected by the input ordering. Therefore to quantify the sensitivity of a model to changes in the input order, one can measure the expected divergence $\mathcal{D}$ between the distributions resulting from any two possible permutations $\mathrm{A}_{\sigma_j}$, $\mathrm{A}_{\sigma_m}$:
\begin{eqnarray}
    \label{eq:pairwise-bias}
   \lefteqn{{\tt ps}(x,{\cal A}) = }\\
    &&
    \mathbbm{E}_{\sigma_{j, m}}
    \left[
    \mathcal{D}
    \Big[
    {\tt P}\hspace{-0.6mm}\left(. \vert x, \mathrm{A}_{\sigma_j}\right) ;
    {\tt P}\hspace{-0.6mm}\left(. \vert x, \mathrm{A}_{\sigma_m}\right)
    \Big]
    \right]
    \nonumber
\end{eqnarray}
%
%

\vspace{2mm}\noindent \textbf{Positional Bias}. A possible cause for permutation sensitivity may be systematic bias, where the most obvious form of bias would be a global preference for a specific option. To measure if there is any systematic preference for certain labels irrespective of the option, one can alternatively look at the average probability mass associated with each option label $w_k$ over all permutations:
%
%
\begin{equation}
    {\tt P}_{\sigma}(w_k|x, \mathcal{A}) = \mathbbm{E}_{\sigma} \! \left[ {\tt P}\hspace{-0.6mm}\left(w_k|x, \mathrm{A}_{\sigma}\right) \right]
\end{equation}
%
%
Note that this marginalized distribution looks at the probability of the $k$-th option irrespective of how the answers have been presented.
If this \textit{positional distribution} is non-uniform ${\tt U}(\cdot)$, the natural interpretation is that the underlying LLM has a biased preference for a particular position. Therefore, a measure of \textit{positional bias} can be defined as the divergence between the positional and uniform distribution 
\begin{equation}
    {\tt pb}(x,{\cal A}) = \mathcal{D}[{\tt P}_{\sigma}(.|x, \mathcal{A}) ; {\tt U}(\cdot)]
\end{equation}
Note that positional bias is more relaxed than permutation sensitivity; a system that is permutation insensitive guarantees having no positional bias, while the reverse is not true.

\subsection{Debiasing Approaches}
To minimize the permutation sensitivity and/or positional bias, we consider two different debiasing strategies that by design enforce invariance.

\vspace{2mm}\noindent \textbf{Permutation debiasing.} A simple approach for correcting permutation sensitivity is to ensemble all permutations
\begin{equation}
    {\tt P}(a_i|x, \mathcal{A}) = \mathbbm{E}_{\sigma} \! \left[ {\tt P}\hspace{-0.6mm} \left(a_i|x, \mathrm{A}_{\sigma}\right) \right]
\end{equation}
This approach eliminates any permutation sensitivity and therefore by definition, positional bias. However, it would require $K!$ passes through the LLM which could be prohibitively expensive. Approximate approaches such as cyclic permutations \cite{zheng2023large} can be used, but they still require $K$ passes and are also computationally expensive at inference time.

\vspace{2mm}\noindent \textbf{Prior-matching}. Instead of cycling through all possible permutations and correcting for permutation sensitivity, a simpler alternative is to focus on minimizing positional bias. Consider introducing a set of weights $\bm{\alpha} = \alpha_{1:K} \in \mathbbm{R}_{+}^{K}$ to scale the original LLM probabilities associated with each particular option label:
%
%
\begin{equation*}
    {\tt P}(w_k \vert x, \mathrm{A}_{\sigma}, \bm\alpha) \!=\! \frac{\alpha_k{\tt P}_{\tt LLM}(w_k|{\tt t}(x, \mathrm{A}_{\sigma}))}{\sum_{j} \! \alpha_j{\tt P}_{\tt LLM}(w_j \vert {\tt t}(x, \mathrm{A}_{\sigma}))}
\end{equation*}
One can then find the weights $\bar{\bm{\alpha}}$ that ensure the system has minimal positional bias\footnote{One can alternatively find $\bm{\alpha}$ to minimize permutation sensitivity, but initial results yielded similar performance to prior-matching} \cite{liusie2023mitigating, zhao2021calibrate} and that the prior over positions is uniform over all questions. 
\begin{equation}
    \bar{\bm{\alpha}} = \underset{\bm{\alpha}}{\text{argmin}}\sum_k \Big{|} {\tt P}_{\sigma}(w_k|x, \mathcal{A}, \bm{\alpha}) - \frac{1}{K}\Big{|}
\end{equation}

\section{Teacher-Student Training for Debiasing}

To address the computational inefficiencies linked with permutation debiasing, this section proposes using teacher-student training to investigate if a smaller inference efficient proxy system ${\tt P}_{\theta}$ could emulate the characteristics of the debiased teacher distribution. Instead of performing $K!$ calls to obtain a permutation debiased prediction, a proxy student could potentially achieve it in a single call.
Our approach is general and is applicable to both white and black-box systems, without the need for labelled data. Although we focus on correcting permutation sensitivity, the framework can be applied for any task invariance and debiasing strategy. 

\subsection{Distillation}
\label{ssec:distillation}

The most inference-efficient approach is to knowledge distil the debiased teacher distribution onto a small non-autoregressive student ${\tt P}_{\theta}$. Given the input $x$ and ordered options $\mathrm{A}_{\sigma}$, the student can be designed to model the debiased teacher distribution,
\begin{align}
    {\tt P}_{\theta}(a | x, \mathrm{A}_{\sigma}) \approx 
    {\tt P}(a | x, \mathcal{A}) \quad \forall \{x, \mathcal{A}\}, \sigma
\end{align}
That is, irrespective of how the possible answers are presented to the student, it should predict consistent distributions that agree with the debiased teacher. This is achieved by minimizing the KL-divergence \cite{distillationhinton}:
\begin{align*}
    \label{eq:cross-entropy}
    \mathcal{L}(\theta) = 
    \mathbbm{E}_{\{x, \mathcal{A}\}, \sigma} 
    \Big[ {\tt KL}\big(
    {\tt P}(\cdot|x, \mathcal{A}) || 
    {\tt P}_{\theta}(\cdot|x, \mathrm{A}_{\sigma}) \big)\Big]
\end{align*}
During training, the debiased teacher probabilities still have to be computed which requires $K!$ white-box calls for every single data point. However, once the student has been trained it can be used independently of the original LLM, and be significantly faster. Since white-box access is not guaranteed, Section \ref{ssec:black-box} discusses how to apply teacher-student training to black-box settings.


\subsection{Error Correction}
\label{ssec:errorc}

For complex tasks, the capacity of a small proxy system might be insufficient. Instead of tasking a student with directly performing the task as in the section above, we consider an error correction student. In addition to the task information and a permutation of the answers, the student receives a sample from a biased teacher, with the aim of emulating the debiased teacher distribution and possibly correcting the initial biased sample. 
%
%
\begin{align}
    \tilde{a} &\sim {\tt P}(a_k | x, \mathrm{A}_{\sigma}) \\
    {\tt P}_{\theta}(a_k | x, \mathrm{A}_{\sigma}, \tilde{a})  &\approx  {\tt P}(a_k | x, \mathcal{A})
\end{align}
Similarly to distillation, the student can be trained by minimising the KL divergence between the proxy and the debiased teacher:
\begin{align*}
    \label{eq:ec-cross-entropy}
    \mathcal{L}(\theta) = 
    \mathbbm{E}_{\{x, \mathcal{A}\}, \sigma} 
    \bigg[
    \mathbbm{E}_{\tilde{a} \sim {\tt P}(\cdot | x, \mathrm{A}_{\sigma})}
    \Big[ \hspace{20mm} \\
    {\tt KL}\big(
    {\tt P}(\cdot|x, \mathcal{A}) || 
    {\tt P}_{\theta}(\cdot|x, \mathrm{A}_{\sigma}, \tilde{a})
    \big)
    \Big]
    \bigg]
\end{align*}
At inference time this model requires a single biased black-box sample from the LLM to produce an approximation to the full debiased distribution of the teacher.

\subsection{Black-Box Considerations} 
\label{ssec:black-box}

The approaches outlined in Sections \ref{ssec:distillation} \& \ref{ssec:errorc} have assumed white-box access to the debiased teacher distribution ${\tt P}(a | x, \mathcal{A})$ during training. In black-box settings, this is not true and a hierarchical monte-carlo approximation to the debiased teacher needs to be used:
\begin{align}
    {\sigma} &\sim \{ {\sigma_1}, {\sigma_2}, ..., {\sigma_{K!}} \} \\
    \tilde{a}^{(i)} &\sim {\tt P}(a | x, \mathrm{A}_{\sigma})
\end{align}
where a random permutation of answers $\mathrm{A}_{\sigma}$ is first chosen followed by sampling from the resulting biased distribution. In expectation, we regain the debiased distribution and can therefore use a sample-based approximation:
\begin{eqnarray*}
{\tt P}(a | x, \mathcal{A}) 
\!\!\!\!&=&\!\!\!\!
\mathbbm{E}_{\sigma} \left[ {\tt P}(a | x, \mathrm{A}_{\sigma}) \right]  \\
    \!\!\!\!&=&\!\!\!\!
    \mathbbm{E}_{\sigma}\left[\mathbbm{E}_{\tilde{a}} \left[ \mathbbm{1}(\tilde{a}=a|x,\mathrm{A}_\sigma) \right] \right]\\
    \!\!\!\!&\approx&\!\!\!\!
    \frac{1}{N} \sum_i  \mathbbm{1}(\tilde{a}^{(i)}=a)\\
\end{eqnarray*}
%
%
Furthermore, the monte-carlo approximation for the knowledge distillation criteria becomes:
\begin{align*}
    \label{eq:mc-cross-entropy}
    \mathcal{L}(\theta) 
    & = \mathbbm{E}_{\{x, \mathcal{A}\}, \sigma} 
    \Big[ {\tt KL}\big(
    {\tt P}(\cdot|x, \mathcal{A}) || 
    {\tt P}_{\theta}(\cdot|x, \mathrm{A}_{\sigma}) \big)\Big] \\
    & \stackrel{c}{=} \mathbbm{E}_{\{x, \mathcal{A}\}, \sigma} \Big[ \mathbbm{E}_{{\tt P}(\cdot|x, \mathcal{A})} \big[ -\ln{\tt P}_{\theta}(\cdot|x, \mathrm{A}_{\sigma}) \big] \Big] \\
    & \approx \mathbbm{E}_{\{x, \mathcal{A}\}, \sigma} \Big[ \frac{1}{N} \sum_i -\ln{\tt P}_{\theta}(\tilde{a}^{(i)}|x, \mathrm{A}_{\sigma}) \Big]
\end{align*}
This allows us to train student models that can emulate the debiased teacher distribution without white-box access to the original LLM. Alternative divergence-based loss functions such as Reverse KL would not cleanly decompose into a black-box compatible form. Note, that for error correction students, an extra LLM sample is required as input to the student.

\section{Experimental Set Up}
\subsection{Datasets}
Experiments are done on two forms of tasks: Multiple Choice Question Answering (MCQA) and Comparative Assessment. For MCQA, we utilize three popular datasets: \textbf{RACE++} \cite{lai-etal-2017-race, pmlr-v101-liang19a}, which consists of English comprehension questions designed for Chinese students spanning from middle school to college. \textbf{CosmosQA} \cite{huang-etal-2019-cosmos}, a large-scale commonsense-based reading comprehension dataset of passages and questions assessing comprehension. \textbf{ARC-CHALLENGE} \cite{clark2018think}, which contain challenging science exam questions drawn from a variety of sources. All datasets have (or are filtered to) 4 options per question. 

For comparative assessment, \textbf{SummEval} \cite{fabbri2020summeval} is used. SummEval is a summary evaluation benchmark of 100 passages and 16 machine-generated summaries per passage, where human annotators have evaluated each summary on coherency (COH), consistency (CON), fluency (FLU), and relevancy (REL). We use the first 70 passages for training, the next 10 for validation, and the final 20 for evaluation.

\subsection{Base Language Models}
Two different open-sourced LLM families are investigated in this work for their general task-solving abilities: FlanT5 \cite{chung2022scaling}, which is a seq2seq T5 \cite{raffel2020exploring} system that has been further instruction tuned on a diverse set of 1600+ NLP tasks \cite{wang-etal-2022-super}; and Llama2-chat \cite{touvron2023llama}, which is a decoder-only language model that is further fine-tuned and optimized for dialogue use cases. A range of the model sizes are considered: 3B and 11B for FlanT5, and 7B and 13B for Llama2-chat.

\begin{table*}[t]
  \centering \footnotesize
   \renewcommand\tabcolsep{4pt}
   \begin{tabular}{|l|cc|cc|cc||cc|cc|cc|cc|}
        \toprule
        &  \multicolumn{6}{c||}{MCQA} &  \multicolumn{8}{c|}{SummEval} \\
        \midrule
        & \multicolumn{2}{c|}{RACE++} & \multicolumn{2}{c|}{COSMOS} & \multicolumn{2}{c||}{ARC-CHAL} &
        \multicolumn{2}{c|}{COH} & \multicolumn{2}{c|}{CON} & \multicolumn{2}{c|}{FLU}   & \multicolumn{2}{c|}{REL}  \\
        & {\tt acc} & {\tt ps} & {\tt acc} & {\tt ps} & {\tt acc} & {\tt ps}  
        & {\tt acc} & {\tt ps} & {\tt acc} & {\tt ps} & {\tt acc} & {\tt ps} & {\tt acc} & {\tt ps} \\
        \midrule
        \rowcolor{Gray} 
        \multicolumn{5}{c}{FlanT5-3B} & \multicolumn{10}{c}{} \\
        baseline (biased)  & 86.7 & 0.09 & 85.7 & 0.12 & 73.6 & 0.22 & 69.2 & 0.20 & 81.1 & 0.12 & 62.5 & 0.17 & 64.5 & 0.17 \\
        prior-matching     & 86.5 & 0.12 & 85.6 & 0.12 & 73.0 & 0.21 & 70.2 & 0.15 & 80.6 & 0.12 & 64.3 & 0.13 & 64.3 & 0.15 \\
        ctx prior-matching & 86.7 & 0.07 & 86.0 & 0.09 & 73.9 & 0.14 & 70.9 & 0.12 & 80.8 & 0.11 & 64.4 & 0.12 & 64.5 & 0.12 \\
        perm-debias        & 87.3 & 0.00 & 86.1 & 0.00 & 74.1 & 0.00 & 71.7 & 0.00 & 82.0 & 0.00 & 65.7 & 0.00 & 65.4 & 0.00 \\
        \midrule         
        \rowcolor{Gray} 
        \multicolumn{5}{c}{FlanT5-11B} & \multicolumn{10}{c}{} \\
        baseline (biased)  & 88.8 & 0.10 & 85.8 & 0.14 & 76.7 & 0.21 & 61.6 & 0.42 & 70.5 & 0.38 & 55.6 & 0.44 & 62.8 & 0.39 \\
        prior-matching     & 88.3 & 0.11 & 86.0 & 0.12 & 76.8 & 0.20 & 67.2 & 0.16 & 77.8 & 0.14 & 58.9 & 0.16 & 64.8 & 0.15 \\
        ctx prior-matching & 88.8 & 0.06 & 86.5 & 0.09 & 77.9 & 0.13 & 67.8 & 0.13 & 77.7 & 0.12 & 59.2 & 0.12 & 65.6 & 0.13 \\
        perm-debias        & 88.9 & 0.00 & 87.4 & 0.00 & 77.9 & 0.00 & 68.9 & 0.00 & 79.7 & 0.00 & 61.4 & 0.00 & 67.0 & 0.00 \\
        \midrule         
        \rowcolor{Gray} 
        \multicolumn{5}{c}{Llama-7B} & \multicolumn{10}{c}{} \\
        baseline (biased)  & 61.2 & 0.67 & 62.1 & 0.65 & 58.5 & 0.67 & 62.8 & 0.30 & 64.1 & 0.49 & 58.0 & 0.42 & 58.3 & 0.56 \\
        prior-matching     & 61.9 & 0.58 & 64.0 & 0.57 & 58.5 & 0.63 & 62.5 & 0.29 & 62.8 & 0.28 & 59.5 & 0.30 & 62.1 & 0.28 \\
        ctx prior-matching & 66.7 & 0.35 & 67.6 & 0.41 & 62.3 & 0.40 & 63.7 & 0.19 & 64.2 & 0.17 & 59.9 & 0.20 & 63.4 & 0.18 \\
        perm-debias        & 68.3 & 0.00 & 72.0 & 0.00 & 64.3 & 0.00 & 64.8 & 0.00 & 66.2 & 0.00 & 59.7 & 0.00 & 65.7 & 0.00 \\
        \midrule         
        \rowcolor{Gray} 
        \multicolumn{5}{c}{Llama-13B} & \multicolumn{10}{c}{} \\
        baseline (biased)  & 71.3 & 0.43 & 68.1 & 0.51 & 68.8 & 0.47 & 62.3 & 0.38 & 71.8 & 0.23 & 58.9 & 0.56 & 63.8 & 0.40 \\
        prior-matching     & 71.8 & 0.39 & 68.7 & 0.45 & 69.0 & 0.46 & 66.0 & 0.22 & 72.6 & 0.17 & 61.7 & 0.25 & 65.5 & 0.24 \\
        ctx prior-matching & 73.3 & 0.25 & 70.9 & 0.35 & 70.3 & 0.31 & 66.7 & 0.16 & 72.1 & 0.14 & 62.4 & 0.18 & 65.7 & 0.17 \\
        perm-debias        & 74.6 & 0.00 & 75.0 & 0.00 & 70.6 & 0.00 & 68.6 & 0.00 & 73.1 & 0.00 & 63.5 & 0.00 & 66.3 & 0.00 \\ 
        \bottomrule
        \end{tabular}
    \caption{Accuracy ({\tt acc}) and permutation sensitivity ({\tt ps}, \S \ref{sec:bias_metrics}) for various LLMs when prompted for MCQA or for pairwise comparative assessment. Llama2-7B is used as the teacher for RACE++ and FlanT5-11b for SummEval. 'ctx- prior matching' refers to applying prior matching to each input over all permutations.}
    \label{tab:bias_and_perf}
    \vspace{-3mm}
\end{table*}

\begin{table*}[t]
    \centering 
    \small
    \renewcommand\tabcolsep{6pt}
    \begin{tabular}{ll||cc||cc|cc|cc|cc}
        \toprule
        & & & & \multicolumn{8}{|c}{SummEval}  \\
        \midrule
        & & \multicolumn{2}{c||}{RACE++} & \multicolumn{2}{c|}{COH} & \multicolumn{2}{c|}{CON} & \multicolumn{2}{c|}{FLU} & \multicolumn{2}{c}{REL} \\
        & type & {\tt acc} & {\tt ps} & {\tt acc} & {\tt ps} & {\tt acc} & {\tt ps} & {\tt acc} & {\tt ps} & {\tt acc} & {\tt ps} \\
        \midrule
        \rowcolor{Gray} 
        \multicolumn{1}{c}{Teachers} & \multicolumn{11}{c}{} \\
        debiased white-box        & & 68.3 & 0.00 & 68.9 & 0.00 & 79.7 & 0.00 & 61.4 & 0.00 & 67.0 & 0.00 \\
        biased white-box          & & 61.2 & 0.67 & 61.6 & 0.42 & 70.5 & 0.38 & 55.6 & 0.44 & 62.8 & 0.39 \\
        expected biased black-box & & 58.4 & - & 58.5 & - & 65.5 & - & 54.3 & - & 58.8 & -\\
        \midrule
        \rowcolor{Gray} 
        \multicolumn{1}{c}{Students} & \multicolumn{11}{c}{} \\
        RoBERTa-base (110M) & {\tt d}   & 26.7 & 0.07 & 61.5 & 0.05 & 70.5 & 0.07 & 61.2 & 0.05 & 60.9 & 0.06 \\
        RoBERTa-base (110M) & {\tt ec}  & 61.4 & 0.37 & 66.4 & 0.04 & 71.7 & 0.06 & 61.6 & 0.03 & 61.9 & 0.05 \\
        \midrule
        DeBERTa-base (110M) & {\tt d}   & 26.9 & 0.05 & 62.6 & 0.03 & 67.1 & 0.04 & 62.1 & 0.03 & 63.0 & 0.05 \\
        DeBERTa-base (110M) & {\tt ec}  & 64.1 & 0.31 & 66.0 & 0.03 & 71.1 & 0.04 & 64.1 & 0.03 & 62.1 & 0.06 \\
        \midrule
        RoBERTa-large (330M) & {\tt d}  & 26.9 & 0.09 & 64.8 & 0.05 & 67.6 & 0.06 & 62.7 & 0.05 & 62.1 & 0.05\\
        RoBERTa-large (330M) & {\tt ec} & 68.0 & 0.25& 66.7 & 0.05 & 72.0 & 0.05 & 63.3 & 0.04 & 63.6 & 0.05 \\
        \midrule
        DeBERTa-large (330M) & {\tt d}  & 47.9 & 0.11 & 65.1 & 0.04 & 71.5 & 0.04 & 64.9 & 0.03 & 63.2 & 0.03\\
        DeBERTa-large (330M) & {\tt ec} & 68.1 & 0.25 & 66.1 & 0.03 & 70.9 & 0.02 & 64.8 & 0.03 & 63.3 & 0.04\\
        \bottomrule
    \end{tabular}
    \caption{Performance of a student trained to emulate the debiased teacher, measured with task accuracy ({\tt acc}) and permutation sensitivity ({\tt ps}). The students are either directly distilled ({\tt d}, \S \ref{ssec:distillation}) or trained to correct the distribution of a single biased black-box teacher decision ({\tt ec}, \S \ref{ssec:distillation}).}
    \label{tab:distillation_results}
\end{table*}

\subsection{Proxy models}
For the student proxy models, only simple encoder-only models are considered. We consider both RoBERTa \cite{liu2019roberta} and DeBERTa-v3 \cite{he2020deberta}, where both the base (110M) and large (330M) size are investigated. The input to the student system proxy is matched to that of the teacher, however for error correction, we further provide the biased teacher decision by appending text to the end of the input prompt. E.g. If the sampled biased teacher prediction was {"A"}, then we concatenate \texttt{Prediction:\;A} to the end of the input text.

\subsection{Methodology}
When applying teacher-student training for debiasing, the debiased white-box teacher distributions are used to train the student. We train 4 seeds per RACE++ setting and 6 seeds per SummEval setting and report the average performance and average sensitivity. For RACE++ Llama2-7b is used as the teacher, while for SummEval FlanT5-11b is used as the teacher. Details of the hyperparameters can be found in appendix \ref{sec:hyperparameters}. 

For each task, we provide the performance of the teacher under different settings. \textbf{Debiased white-box teacher} refers to the performance when permutation debiased decisions are used. \textbf{Biased white-box teacher} performance refers when the prediction is taken as the argmax of a single teacher call. The \textbf{expected biased black-box performance} is the expected accuracy when samples from the teacher are drawn from the underlying biased distribution. Note that accuracy may differ from the biased white-box accuracy, if the decisions are not well calibrated \cite{guo2017calibration}. 
When evaluating permutation sensitivity, total variation is used since the KL divergence is unbounded and, if used, metrics may be overly influenced by individual samples that largely diverge.

\section{Results}
\subsection{Permutation Bias of LLMs}
Table \ref{tab:bias_and_perf} shows the performance and permutation sensitivity for various LLMs on a range of multiple choice answering tasks, as well as for comparative assessment, and demonstrates the following points:

\textbf{1) LLMs may fail to adhere to task invariances.} Both Llama2 and FlanT5 style models exhibit high permutation sensitivity across various tasks. Llama2, in particular, shows reasonable accuracy across a range of tasks, however also has high permutation sensitivity in nearly all tasks. This highlights that the output distribution of prompted LLMs can be largely influenced by the order of the input options. 

\textbf{2) Models that satisfy positional invariance for some tasks, may not be positional invariant for all tasks.} FlanT5-3B and FlanT5-11B demonstrate minimal permutation sensitivity for all MCQA tasks, likely due to the additional fine-tuning of FlanT5 on a variety of tasks including multiple choice question answering exams. This fine-tuning has likely imparted implicit permutation invariance for tasks resembling those encountered during training. However, when FlanT5-11B is applied to comparative assessment, the system exhibits considerable permutation sensitivity across all attributes of SummEval. This implies that further training on supervised data may mitigate bias and implicitly impart invariances, however, such a solution is task-specific and may not necessarily generalize to tasks seen beyond training. 

\textbf{3) Addressing neglected invariances can yield significantly better task performance.} Permutation debiasing guarantees zero permutation sensitivity, and applying the method can yield large improvements in performance for many tasks. Even tasks with low permutation sensitivity (e.g. FlanT5 on MCQA) gain small performance boosts, though in settings with high bias one can gain up to 10\% in accuracy. Further, a loose correlation between permutation sensitivity and accuracy can be observed across tasks and models. 

\textbf{4) Positional Bias alone does not account for the observed positional bias.} Applying prior matching, which ensures that there is no positional bias towards any of the label tokens, alone does not resolve the permutation sensitivity. Although in some cases this can significantly improve both sensitivity and accuracy (e.g. FlanT5-11B comparative assessment), for some tasks, permutation sensitivity may remain significant and performance can be substantially worse than permutation debiasing. 

\textbf{5) Context Positional Bias can account for much of the observed performance degradation.} As an extension to prior matching, we also consider context-prior matching where prior matching is applied over all $K!$ permutations of the particular input. This enables one to capture the positional bias caused by the specific input prompt. Correcting for this bias yields performance closely matching that of permutation debiasing, highlighting that a positional bias can exist for particular contexts. However note that, unlike prior matching, context-prior matching requires $K!$ calls and is only useful as analysis relative to permutation sensitivity.  

\subsection{Debiased Student Performance}
Table \ref{tab:distillation_results} shows the performance of various students when trained to emulate the teacher debiased decisions, where students are either purely distilled (\S \ref{ssec:distillation}) or trained to achieve error-correction (\S \ref{ssec:errorc}). The table shows that:

\textbf{1)} For some tasks (e.g. comparative assessment on SummEval) the teacher's abilities can be adequately learned by a smaller student through standard knowledge distillation. The resulting student can achieve performance considerably better than the biased teacher and low permutation sensitivity, all while being considerably more computationally efficient.

\textbf{2)} For complex tasks (e.g. RACE++) the student is not powerful enough to alone capture the abilities of the teacher. However, in such cases, error correction students can effectively leverage a single-biased teacher decision to predict the estimated general debiased distributions. These student systems are more robust to changes in permutations, although are not fully permutation invariant. Note that error correction consistently yields better performance than copying the biased teacher's decision, illustrating that the students can capture useful information of the underlying teacher's prediction space. 

\textbf{3)} Although the size and ability of student can be an important factor when applying the framework (e.g. RACE++), for some tasks the required model complexity can saturate early and a further increase in size/ability does not impact downstream performance.

\subsection{Black-Box Training Efficiency}
\begin{figure}[H]
    \centering
    \includegraphics[width=\columnwidth]{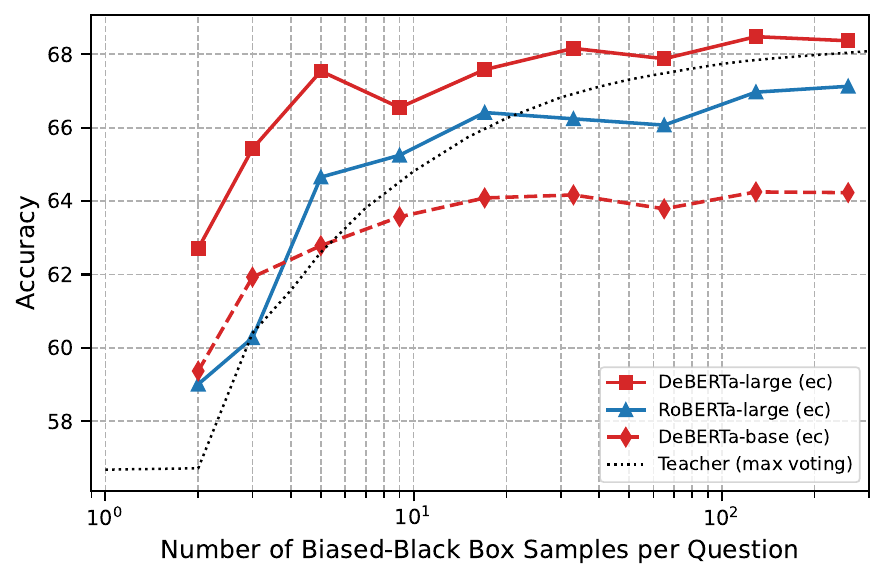}
    \caption{RACE++ performance of error correction students when using $N$ black-box samples to approximate the debiased distribution (\S \ref{ssec:black-box})}
\label{fig:RACE++_sample_efficiency}
\end{figure}

\noindent The previous section applied the teacher-student training framework assuming white-box access during training. Although infinite black-box samples can be used to derive the underlying distribution, this section investigates the sample efficiency of the framework in black-box settings. Figure \ref{fig:RACE++_sample_efficiency} displays the RACE++ performance of an error correction student when trained using $N$ black-box teacher samples per example. The curve illustrates that teacher-student training does not require an excessive number of black-box samples, with performance saturating at 32 samples per example. Interestingly, when using only a few samples, DeBERTa-large can outperform the max-voting performance of the debiased teacher. This implies that by applying teacher-student training, the student can infer the systematic biases present in the teacher, and yield corrected distributions from many noisy approximations. The analysis was done for RACE++, and as having more options would require more samples to approximate the true underlying distribution, one would expect comparative assessment to require fewer black-box samples per input.

\subsection{Impact of Data Size}
\begin{figure}[H]
    \centering
    \includegraphics[width=\columnwidth]{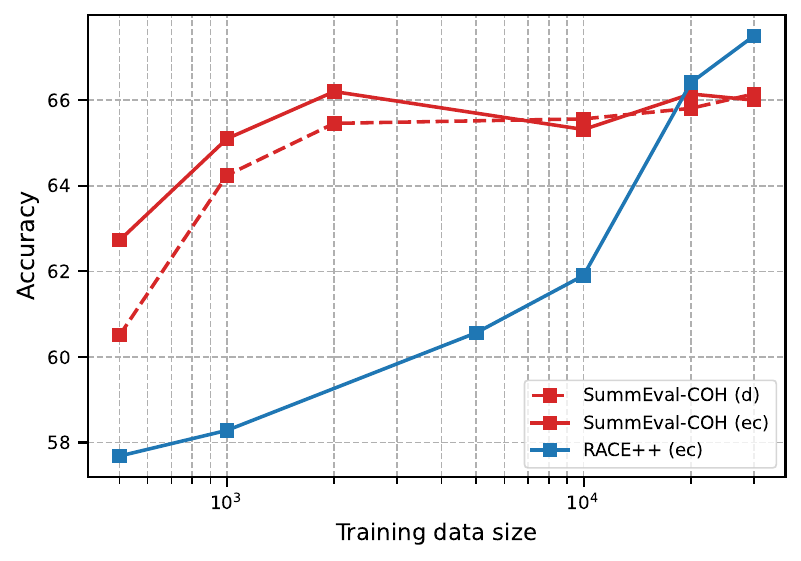}
    \caption{DeBERTa-large accuracy when using a limited number of examples during teacher-student training.}
    \label{fig:training_efficiency}
\end{figure}

Figure \ref{fig:training_efficiency} shows the effectiveness of a student DeBERTa-large model when trained on a limited number of training samples. The plot shows that the number of samples required before performance saturates varies largely on task complexity. For comparative assessment on SummEval coherency, only 2000 examples are required, while RACE++ requires 30,000 examples before a DeBERTa-large error correction student reaches the debiased teacher performance on RACE++. 

\section{Conclusions}
This paper explores the sensitivity of LLMs to the order of input options for multiple-choice question-answering and comparative assessment. We illustrate the effectiveness of various debiasing approaches for mitigating these biases and the associated performance improvement. While these debiasing methods often entail high computational costs, we show that teacher-student training can yield inference-efficient student models capable of emulating a debiased teacher distribution. Our approach is practical in both white-box and black-box settings, requiring a manageable number of training data points and black-box samples.

\section{Limitations}
The teacher-student training was demonstrated to be effective for multiple choice question answering and comparative assessment, however was not demonstrated to invariances beyond permutation sensitivity. Further, in the current framework, the training domain matches the downstream evaluation domain. Though this is a useful set-up for some scenarios, it does not investigate cross-task generalization or whether the students generalize to tasks that differ mildly from those in training. Our work also currently requires access to unlabelled input examples, which the teacher then produces predictions for.

\section{Acknowledgements}
This paper reports on research supported by Cambridge University Press \& Assessment (CUP\&A), a department of The Chancellor, Masters, and Scholars of the University of Cambridge. This research is further supported by the Gates Cambridge Trust.

\vspace{-5mm}

                    

\bibliography{anthology,custom}
\bibliographystyle{acl_natbib}

\appendix

\onecolumn
\section{Hyperparameter Settings}
\label{sec:hyperparameters}
We train over the entire training dataset with 2 epochs, with a batch size of 4 and learning rate of 1e-5 for the base students and 5e-6 for the large students, using the AdamW optimizer. The learning rate was selected through a 1D search using 3 seeds for SummEval-COH, with the best learning rate in the set $\{1e^{-6}, 2e^{-6}, 5e^{-6}, 1e^{-5}, 2e^{-5}, 5e^{-5}\}$. This learning rate was later kept for all later experiments. Validation is run every 1000 examples, and the checkpoint with best validation accuracy is used at evaluation. Experiments were run on Nvidia L40 GPUs with 50Gb of RAM. For DeBERTa-large Each Summeval seed took 1 hour to run, and each RACE++ seed took 2 hours. 

\section{Further details on LLM Set Up}
For comparative assessment, the label words $w_k$ used are "Summary A" and "Summary B". This is equivalent to appending "Summary" to the end of the input and then calculating the probability of "A" or "B". For llama2-chat, a further "Answer:" is appended to the prompt so the model knows where the input ends and the generated answer ends. 

\section{LLM performance on extended tasks}
Tables \ref{tab:detailed_perm_sens_comp} and \ref{tab:detailed_perm_sens_mcqa} show the LLM performance on further multiple choice and comparative assessment tasks, while also presenting the positional bias observed for the systems and debiasing approaches. Similar trends to those in the main paper are observed over a wider range of tasks.

\section{Detailed Student Performance}
Table \ref{tab:std_table} shows the standard deviations observed in the accuracies for the various student models. We further run experiments on BERT \cite{devlin-etal-2019-bert} and BERT-tiny \cite{jiao2020tinybert} to investigate the ability of weaker students. As expected, the BERT students were observed to be much weaker than their more modern counterparts (RoBERTa and DeBERTa) of equivalent size.

\begin{table}[h]
  \centering \fontsize{7}{10}\selectfont
   \renewcommand\tabcolsep{2pt}
   \begin{tabular}{|l|ccc|ccc|ccc|ccc||ccc|ccc|ccc|ccc|}
        \toprule
        &  \multicolumn{12}{c||}{SummEval} &  \multicolumn{12}{c|}{TopicalChat}\\
        \midrule
                           &  \multicolumn{3}{c|}{COH} & \multicolumn{3}{c|}{CON} & \multicolumn{3}{c|}{FLU}   & \multicolumn{3}{c||}{REL}  & 
                           \multicolumn{3}{c|}{COH} & \multicolumn{3}{c|}{CNT} & \multicolumn{3}{c|}{ENG}   & \multicolumn{3}{c|}{NAT} \\
        & {\tt acc} & {\tt pb} & {\tt ps} & {\tt acc} & {\tt pb} & {\tt ps} & {\tt acc} & {\tt pb} & {\tt ps} & {\tt acc} & {\tt pb} & {\tt ps} & {\tt acc} & {\tt pb} & {\tt ps} & {\tt acc} & {\tt pb} & {\tt ps} & {\tt acc} & {\tt pb} & {\tt ps} & {\tt acc} & {\tt pb} & {\tt ps}\\
        \midrule
        \rowcolor{Gray} 
        \multicolumn{5}{c}{FlanT5-3B} & \multicolumn{20}{c}{} \\
        baseline           & 69.2 & 0.16 & 0.20 & 81.1 & 0.06 & 0.12 & 62.5 & 0.12 & 0.17 & 64.5 & 0.10 & 0.17 & 75.8 & 0.04 & 0.11 & 70.7 & 0.08 & 0.13 & 65.9 & 0.01 & 0.11 & 70.1 & 0.02 & 0.11 \\
        prior-matching     & 70.5 & 0.03 & 0.15 & 80.5 & 0.03 & 0.11 & 64.3 & 0.03 & 0.13 & 64.1 & 0.02 & 0.15 & 75.7 & 0.02 & 0.11 & 71.8 & 0.00 & 0.11 & 65.5 & 0.02 & 0.11 & 69.5 & 0.06 & 0.12 \\
        ctx prior-matching & 70.9 & 0.00 & 0.12 & 80.8 & 0.01 & 0.11 & 64.4 & 0.01 & 0.12 & 64.5 & 0.01 & 0.12 & 75.4 & 0.01 & 0.09 & 71.6 & 0.00 & 0.09 & 66.1 & 0.00 & 0.09 & 69.4 & 0.00 & 0.10 \\
        perm-debias        & 71.7 & 0.00 & 0.00 & 82.0 & 0.00 & 0.00 & 65.7 & 0.00 & 0.00 & 65.4 & 0.00 & 0.00 & 76.2 & 0.00 & 0.00 & 72.6 & 0.00 & 0.00 & 66.6 & 0.00 & 0.00 & 70.9 & 0.00 & 0.00 \\
        \midrule         
        \rowcolor{Gray} 
        \multicolumn{5}{c}{FlanT5-11B} & \multicolumn{20}{c}{} \\
        baseline           & 61.6 & 0.42 & 0.42 & 70.5 & 0.37 & 0.38 & 55.6 & 0.44 & 0.44 & 62.8 & 0.39 & 0.39 & 69.2 & 0.29 & 0.29 & 62.4 & 0.35 & 0.35 & 66.3 & 0.29 & 0.29 & 68.1 & 0.30 & 0.30 \\
        prior-matching     & 67.3 & 0.02 & 0.16 & 77.8 & 0.00 & 0.14 & 58.9 & 0.03 & 0.16 & 65.3 & 0.05 & 0.16 & 74.9 & 0.14 & 0.17 & 75.3 & 0.04 & 0.13 & 73.5 & 0.06 & 0.13 & 73.5 & 0.14 & 0.17 \\
        ctx prior-matching & 67.8 & 0.00 & 0.13 & 77.7 & 0.02 & 0.12 & 59.2 & 0.02 & 0.12 & 65.6 & 0.00 & 0.13 & 77.1 & 0.00 & 0.09 & 75.5 & 0.00 & 0.10 & 73.6 & 0.00 & 0.10 & 74.6 & 0.00 & 0.10 \\
        perm-debias        & 68.9 & 0.00 & 0.00 & 79.7 & 0.00 & 0.00 & 61.4 & 0.00 & 0.00 & 67.0 & 0.00 & 0.00 & 77.9 & 0.00 & 0.00 & 79.6 & 0.00 & 0.00 & 75.0 & 0.00 & 0.00 & 75.8 & 0.00 & 0.00 \\
        \midrule         
        \rowcolor{Gray} 
        \multicolumn{5}{c}{Llama-7B} & \multicolumn{20}{c}{} \\
        baseline           & 62.8 & 0.09 & 0.30 & 64.1 & 0.45 & 0.49 & 58.0 & 0.33 & 0.42 & 58.3 & 0.53 & 0.56 & 63.3 & 0.12 & 0.31 & 60.7 & 0.44 & 0.45 & 60.9 & 0.51 & 0.52 & 60.3 & 0.27 & 0.34 \\
        prior-matching     & 62.5 & 0.07 & 0.29 & 62.8 & 0.13 & 0.29 & 59.6 & 0.10 & 0.30 & 62.0 & 0.01 & 0.27 & 63.3 & 0.12 & 0.31 & 63.0 & 0.04 & 0.27 & 64.5 & 0.05 & 0.30 & 62.0 & 0.07 & 0.31 \\
        ctx prior-matching & 63.7 & 0.00 & 0.19 & 64.2 & 0.01 & 0.17 & 59.9 & 0.03 & 0.20 & 63.4 & 0.00 & 0.18 & 63.2 & 0.02 & 0.19 & 65.1 & 0.01 & 0.18 & 66.3 & 0.00 & 0.19 & 61.4 & 0.02 & 0.22 \\
        perm-debias        & 64.8 & 0.00 & 0.00 & 66.2 & 0.00 & 0.00 & 59.7 & 0.00 & 0.00 & 65.7 & 0.00 & 0.00 & 63.5 & 0.00 & 0.00 & 65.5 & 0.00 & 0.00 & 67.0 & 0.00 & 0.00 & 63.3 & 0.00 & 0.00 \\
        \midrule         
        \rowcolor{Gray} 
        \multicolumn{5}{c}{Llama-13B} & \multicolumn{20}{c}{} \\
        baseline           & 62.3 & 0.36 & 0.38 & 71.8 & 0.18 & 0.23 & 58.9 & 0.56 & 0.56 & 63.8 & 0.38 & 0.40 & 63.4 & 0.29 & 0.33 & 64.5 & 0.37 & 0.40 & 70.9 & 0.28 & 0.31 & 60.1 & 0.35 & 0.39 \\
        prior-matching     & 65.8 & 0.06 & 0.23 & 72.9 & 0.01 & 0.17 & 61.6 & 0.03 & 0.24 & 65.5 & 0.11 & 0.24 & 64.9 & 0.04 & 0.20 & 67.9 & 0.04 & 0.24 & 73.4 & 0.04 & 0.20 & 64.3 & 0.03 & 0.22 \\
        ctx prior-matching & 66.7 & 0.00 & 0.16 & 72.1 & 0.02 & 0.14 & 62.4 & 0.01 & 0.18 & 65.7 & 0.01 & 0.17 & 65.4 & 0.02 & 0.14 & 68.9 & 0.01 & 0.16 & 73.3 & 0.00 & 0.13 & 64.9 & 0.01 & 0.16 \\
        perm-debias        & 68.6 & 0.00 & 0.00 & 73.1 & 0.00 & 0.00 & 63.5 & 0.00 & 0.00 & 66.3 & 0.00 & 0.00 & 67.5 & 0.00 & 0.00 & 70.0 & 0.00 & 0.00 & 74.4 & 0.00 & 0.00 & 65.8 & 0.00 & 0.00 \\ 
        \bottomrule
        \end{tabular}
    \caption{Accuracy ({\tt acc}), permutation bias ({\tt pb}) and permutation sensitivity for various LLMs when prompted for Comparative Assessment.}
    \label{tab:detailed_perm_sens_comp}
\end{table}

\begin{table}[H]
  \centering  \fontsize{8}{10}\selectfont
   \renewcommand\tabcolsep{4pt}
   \begin{tabular}{|l|ccc|ccc|ccc|ccc|ccc|}
        \toprule
        \midrule
        & \multicolumn{3}{c|}{RACE++} & \multicolumn{3}{c|}{COSMOS} & \multicolumn{3}{c|}{ReClor} & \multicolumn{3}{c|}{ARC-EASY} &\multicolumn{3}{c|}{ARC-CHAL} \\
        & {\tt acc} & {\tt pb} & {\tt ps} & {\tt acc} & {\tt pb} & {\tt ps} & {\tt acc} & {\tt pb} & {\tt ps} & {\tt acc} & {\tt pb} & {\tt ps} & {\tt acc} & {\tt pb} & {\tt ps} \\

        \midrule
        \rowcolor{Gray} 
        \multicolumn{5}{c}{FlanT5-3B} & \multicolumn{11}{c}{} \\
        baseline (biased)   & 86.7 & 0.01 & 0.09 & 85.7 & 0.02 & 0.12 & 54.8 & 0.03 & 0.25 & 85.3 & 0.05 & 0.16 & 73.6 & 0.06 & 0.22 \\
        prior-matching      & 86.5 & 0.05 & 0.12 & 85.6 & 0.01 & 0.12 & 54.0 & 0.03 & 0.25 & 85.9 & 0.03 & 0.15 & 73.0 & 0.03 & 0.21 \\
        ctx prior-matching  & 86.7 & 0.00 & 0.07 & 86.0 & 0.00 & 0.09 & 55.4 & 0.00 & 0.17 & 87.0 & 0.00 & 0.09 & 73.9 & 0.00 & 0.14 \\
        perm-debias         & 87.3 & 0.00 & 0.00 & 86.1 & 0.00 & 0.00 & 54.2 & 0.00 & 0.00 & 86.8 & 0.00 & 0.00 & 74.1 & 0.00 & 0.00 \\
        \midrule         
        \rowcolor{Gray} 
        \multicolumn{5}{c}{FlanT5-11B} & \multicolumn{11}{c}{} \\
        baseline (biased)   & 88.8 & 0.03 & 0.10 & 85.8 & 0.05 & 0.14 & 57.0 & 0.10 & 0.30 & 89.5 & 0.05 & 0.14 & 76.7 & 0.06 & 0.21 \\
        prior-matching      & 88.3 & 0.05 & 0.11 & 86.0 & 0.02 & 0.12 & 57.8 & 0.03 & 0.28 & 89.3 & 0.03 & 0.13 & 76.8 & 0.03 & 0.20  \\
        ctx prior-matching  & 88.8 & 0.00 & 0.06 & 86.5 & 0.00 & 0.09 & 58.8 & 0.00 & 0.19 & 90.2 & 0.00 & 0.08 & 77.9 & 0.00 & 0.13 \\
        perm-debias         & 88.9 & 0.00 & 0.00 & 87.4 & 0.00 & 0.00 & 59.6 & 0.00 & 0.00 & 90.2 & 0.00 & 0.00 & 77.9 & 0.00 & 0.00 \\
        \midrule         
        \rowcolor{Gray} 
        \multicolumn{5}{c}{Llama-7B} & \multicolumn{11}{c}{} \\
        baseline (biased)  & 58.1 & 0.32 & 0.72 & 52.2 & 0.30 & 0.73 & 38.8 & 0.61 & 0.99 & 76.2 & 0.15 & 0.45 & 58.5 & 0.24 & 0.67 \\
        prior-matching     & 59.9 & 0.02 & 0.59 & 54.1 & 0.03 & 0.65 & 40.8 & 0.02 & 0.72 & 76.1 & 0.04 & 0.41 & 58.5 & 0.05 & 0.63 \\
        ctx prior-matching & 64.6 & 0.00 & 0.33 & 56.8 & 0.00 & 0.46 & 44.4 & 0.00 & 0.44 & 80.6 & 0.00 & 0.26 & 62.3 & 0.00 & 0.40 \\
        perm-debias        & 66.0 & 0.00 & 0.00 & 60.8 & 0.00 & 0.00 & 48.6 & 0.00 & 0.00 & 83.7 & 0.00 & 0.00 & 64.3 & 0.00 & 0.00 \\
        \midrule         
        \rowcolor{Gray} 
        \multicolumn{5}{c}{Llama-13B} & \multicolumn{11}{c}{} \\
        baseline (biased)   & 71.3 & 0.19 & 0.43 & 63.4 & 0.19 & 0.54 & 49.6 & 0.31 & 0.65 & 82.7 & 0.07 & 0.29 & 68.8 & 0.14 & 0.47 \\
        prior-matching      & 71.8 & 0.05 & 0.39 & 65.1 & 0.01 & 0.48 & 50.2 & 0.06 & 0.59 & 83.1 & 0.02 & 0.28 & 69.0 & 0.05 & 0.46 \\
        ctx prior-matching  & 73.3 & 0.00 & 0.25 & 65.8 & 0.00 & 0.38 & 50.0 & 0.00 & 0.40 & 86.3 & 0.00 & 0.19 & 70.3 & 0.00 & 0.31 \\
        perm-debias         & 74.6 & 0.00 & 0.00 & 70.2 & 0.00 & 0.00 & 53.2 & 0.00 & 0.00 & 87.9 & 0.00 & 0.00 & 70.6 & 0.00 & 0.00 \\ 
        \bottomrule
        \end{tabular}
    \caption{Accuracy ({\tt acc}), permutation bias ({\tt pb}) and permutation sensitivity for various LLMs when prompted for Multiple Choice Question Answering.}
    \label{tab:detailed_perm_sens_mcqa}. 
\end{table}

\begin{table}[H]
    \centering \fontsize{8}{10}\selectfont
    \renewcommand\tabcolsep{10pt}
    \begin{tabular}{l||c||ccccc}
        \toprule
        & &\multicolumn{4}{|c}{SummEval}  \\
        \midrule
        & RACE++ & COH & CON & FLU & REL \\
        \midrule
        \rowcolor{Gray} 
        \multicolumn{1}{c}{Teachers} & \multicolumn{5}{c}{} \\
        debiased white-box        & 68.3 & 68.9 & 79.7 & 61.4 & 67.0 \\
        biased white-box          & 61.2 & 61.6 & 70.5 & 55.6 & 62.8 \\
        expected biased black-box & 58.4 & 58.5 & 65.5 & 54.3 & 58.8 \\
        \midrule
        \rowcolor{Gray} 
        \multicolumn{1}{c}{Distillation} & \multicolumn{5}{c}{} \\
        BERT-tiny (4.4M) & 26.4\std{0.4} & 50.9\std{1.0} & 51.6\std{1.8} & 50.0\std{0.5} & 50.6\std{0.9}  \\
        BERT-base (110M) & 45.6\std{0.2} & 57.7\std{0.7} & 69.5\std{0.6} & 60.9\std{1.2} & 56.8\std{0.4}  \\
        RoBERTa-base (110M) & 26.7\std{0.3} & 61.5\std{5.3} & 70.5\std{1.2} & 61.2\std{1.8} & 60.9\std{0.7}  \\
        DeBERTa-base (110M) & 26.9\std{0.0} & 62.6\std{5.7} & 67.1\std{7.7} & 62.1\std{5.5} & 63.0\std{0.4}  \\
        RoBERTa-large (330M) & 26.9\std{0.0} & 64.8\std{1.2} & 67.6\std{7.7} & 62.7\std{2.9} & 62.1\std{0.5}  \\
        DeBERTa-large (330M) & 47.9\std{21.0} & 65.1\std{0.6} & 71.5\std{0.7} & 64.9\std{0.6} & 63.2\std{0.5}  \\
        \rowcolor{Gray} 
        \midrule
        \multicolumn{1}{c}{Error Correction} & \multicolumn{5}{c}{} \\
        BERT-tiny (4.4M) & 57.7\std{0.6} & 53.9\std{2.0} & 57.5\std{5.1} & 51.6\std{1.6} & 53.0\std{2.4}  \\
        BERT-base (110M) & 58.7\std{0.0} & 57.4\std{0.4} & 70.7\std{0.7} & 61.9\std{1.2} & 58.7\std{0.8}  \\
        RoBERTa-base (110M) & 61.4\std{0.1} & 66.4\std{0.4} & 71.7\std{0.5} & 61.6\std{0.6} & 61.9\std{0.4}  \\
        DeBERTa-base (110M) & 64.1\std{0.1} & 66.0\std{0.5} & 71.1\std{1.3} & 64.1\std{1.0} & 62.1\std{1.5}  \\
        RoBERTa-large (330M) & 68.0\std{0.5}& 66.7\std{0.7} & 72.0\std{1.2} & 63.3\std{0.6} & 63.6\std{0.5}  \\
        DeBERTa-large (330M) & 68.1\std{0.9} & 66.1\std{0.8} & 70.9\std{1.6} & 64.8\std{0.5} & 63.3\std{0.6}  \\
        \bottomrule
    \end{tabular}
    \caption{Results extending the accuracies presented in Table \ref{tab:distillation_results}, providing standard deviations and extended to BERT and BERT-tiny.}
    \label{tab:std_table}
\end{table}

\end{document}